\documentclass[sigconf]{acmart}

\copyrightyear{2026}
\acmYear{2026}
\setcopyright{cc}
\setcctype{by}
\acmConference[SIGIR '26]{Proceedings of the 49th International ACM SIGIR Conference on Research and Development in Information Retrieval}{July 20--24, 2026}{Melbourne, VIC, Australia}
\acmBooktitle{Proceedings of the 49th International ACM SIGIR Conference on Research and Development in Information Retrieval (SIGIR '26), July 20--24, 2026, Melbourne, VIC, Australia}
\acmDOI{10.1145/3805712.3808603}
\acmISBN{979-8-4007-2599-9/2026/07}
\renewcommand{\shortauthors}{Yongan Yu et al.}

\newcommand{\ourmethod}{\textsc{WeatherArchive-Bench}}
\AtBeginDocument{%
  }

\usepackage{multicol}
\usepackage{booktabs}

\usepackage{enumitem}
\usepackage{graphicx}
\usepackage[most]{tcolorbox}
\usepackage{float}
\usepackage{xcolor}
\usepackage{amsmath}
\usepackage{courier}
\usepackage{colortbl}
\usepackage{multirow}
\usepackage{caption}
\usepackage{booktabs}
\usepackage{tabularx}
\usepackage{threeparttable}

\usepackage{amssymb}

\usepackage{xcolor}
\usepackage{pifont}

\newcommand{\cmark}{\textcolor{green!60!black}{\ding{51}}}%
\newcommand{\xmark}{\textcolor{red}{\ding{55}}}%
\begin{document}

\title{WeatherArchive: A Benchmark for Retrieval-Augmented Reasoning over Historical Weather Archives}

\author{Yongan Yu}
\authornote{All authors contributed equally to this study.}
\orcid{0009-0006-3264-1170}
\affiliation{%
  \institution{McGill University}
  \city{Montreal}
  \country{Canada}
}
\email{yongan.yu@mail.mcgill.ca}

\author{Xianda Du}
\authornotemark[1]
\orcid{0009-0006-6500-4020}
\affiliation{%
  \institution{University of Waterloo}
  \city{Waterloo}
  \country{Canada}}
\email{a32du@uwaterloo.ca}

\author{Qingchen Hu}
\authornotemark[1]
\orcid{0009-0000-9008-6757}
\affiliation{%
  \institution{McGill University}
  \city{Montreal}
  \country{Canada}
}
\email{qingchen.hu@mail.mcgill.ca}

\author{Jiahao Liang}
\authornotemark[1]
\orcid{0009-0006-9579-078X}
\affiliation{%
  \institution{McGill University}
  \city{Montreal}
  \country{Canada}
}
\email{jiahao.liang@mail.mcgill.ca}

\author{Jingwei Ni}
\orcid{0009-0008-8988-301X}
\affiliation{%
  \institution{ETH Zurich}
  \city{Zurich}
  \country{Switzerland}}
\email{jingni@ethz.ch}

\author{Dan Qiang}
\orcid{0000-0002-7483-2681}
\affiliation{%
  \institution{McGill University}
  \city{Montreal}
  \country{Canada}}
\email{dan.qiang@mail.mcgill.ca}

\author{Kaiyu Huang}
\authornote{Corresponding Authors}
\orcid{0000-0001-6779-1810}
\affiliation{%
  \institution{Beijing Jiaotong University}
  \city{Beijing}
  \country{China}}
\email{kyhuang@bjtu.edu.cn}

\author{Grant McKenzie}
\orcid{0000-0003-3247-2777}
\affiliation{%
  \institution{McGill University}
  \city{Montreal}
  \country{Canada}}
\email{grant.mckenzie@mcgill.ca}

\author{Renée Sieber}
\authornotemark[2]
\orcid{0000-0002-0720-3498}
\affiliation{%
  \institution{McGill University}
  \city{Montreal}
  \country{Canada}}
\email{renee.sieber@mcgill.ca}

\author{Fengran Mo}
\authornotemark[2]
\orcid{0000-0002-0838-6994}
\affiliation{%
  \institution{Université de Montréal}
  \city{Montreal}
  \country{Canada}}
\email{fengran.mo@umontreal.ca}
\renewcommand{\shortauthors}{Yu et al.}

\begin{abstract}
Historical news segments on weather events are collections of enduring primary source records that offer rich, untapped narratives of how societies have experienced and responded to extreme weather events. 
These qualitative accounts provide insights into societal vulnerability and resilience that are largely absent from meteorological records, making them valuable for climate scientists to understand societal responses. 
However, their large scale, noise in optical character recognition (OCR), and archaic language make it difficult to transform them into structured knowledge for climate research. 
To address this challenge, we introduce \ourmethod, the first large-scale benchmark for evaluating end-to-end retrieval-augmented generation (RAG) systems on historical weather archives.  \textsc{WeatherArchive-Bench} comprises two tasks: \textit{WeatherArchive-Retrieval}, which measures a system’s ability to locate historically relevant news segments from over one million archival news segments, and \textit{WeatherArchive-Assessment}, which evaluates whether Large Language Models (LLMs) can classify societal vulnerability and resilience indicators from extreme weather narratives and answer queries using the segments retrieved.
Extensive experiments across sparse, dense, and re-ranking retrievers, as well as a diverse set of LLMs, reveal that dense retrievers often fail on historical terminology, while LLMs frequently misinterpret vulnerability and resilience concepts. These findings highlight key limitations in reasoning about complex societal indicators and provide insights for designing more robust climate-focused RAG systems from archival contexts. The constructed dataset and evaluation framework are available at: \href{https://github.com/Weather-Archival-Rescue/WeatherArchive-Bench}{https://github.com/Weather-Archival-Rescue/WeatherArchive-Bench}. 
\end{abstract}





\maketitle

\section{Introduction}
\setlength{\tabcolsep}{10pt} 
\begin{table*}[t]
\small
\centering
\caption{Comparison of existing QA benchmarks with \ourmethod{}.}
\begin{tabular}{lccccc}
\toprule
Dataset & \# Papers & Paper Source & Domain & Historical Data & Task \\
\midrule
REPLIQA \citep{monteiro2024repliqa} & 17.9K & Synthetic & General & \xmark & Topic Retrieval + QA \\
CPIQA \citep{mutalik2025cpiqa} & 4.55K & Climate papers & Climate Sci. & \xmark & Multimodal QA \\
ClimRetrieve \citep{schimanski2024climretrieve} & 30 & Reports & Climate Sci. & \xmark & Document Retrieval \\
ClimaQA \citep{manivannan2024climaqa} & 23 & Textbooks & Climate Sci. & \xmark & Scientific QA \\
\rowcolor{gray!20}
WeatherArchive (Ours) & 1.04M & Hist. archives & Climate Sci. & \cmark & Retrieval + QA + classification \\
\bottomrule
\end{tabular}
\label{tab:climate_benchmarks}
\end{table*}
Extreme weather events have become increasingly frequent and severe due to climate change, which results in urgent challenges for climate adaptation and disaster preparedness \citep{o2006climate}. 
Climate policymakers are expected to design targeted adaptation strategies that integrate disaster response with long-horizon planning, including climate-resilient urban development \citep{xu2024resilience} and sustainable land use policies \citep{zuccaro2020future}.
Achieving these goals requires not only meteorological data, but also a deeper understanding of how communities, infrastructures, and economic sectors have responded to climate hazards \citep{bollettino2020public, mallick2024analyzing}.
To this end, historical archives provide such knowledge, documenting past extreme weather events alongside their cascading economic impacts, community responses, and local adaptation practices \citep{yu2025wximpactbench}.
A systematic analysis of these records can reveal which factors were most disruptive during a specified extreme weather event, thereby providing evidence-based insights to inform future climate policy interventions.
However, such resources are largely unavailable to the research community.

As shown in Table~\ref{tab:climate_benchmarks}, existing benchmarks in the climate domain focus on either relatively small-scale paper sources or primarily target scientific papers and reports without historically grounded archival data. 
Furthermore, no standardized evaluation exists for extracting societal vulnerability and resilience indicators from historical records, limiting the development of AI systems that can translate historical climate evidence into actionable policy insights.

Applying RAG systems in this domain is particularly challenging. As such, models combine retrieval and generative reasoning to process large document collections \citep{lewis2020retrieval,mo2025uniconv,mo2026opendecoder}, but historical archives present unique obstacles: OCR errors, archaic terminology, inconsistent formatting, and narratives that intertwine meteorological events with unrelated social or economic commentary \citep{bingham2010digitization, verhoef2015cultural,wang2025finauditing}. These issues impede retrieval of relevant passages and hinder reasoning, as pretrained LLMs often lack exposure to historical terminology and socio-environmental concepts \citep{liu2024datasets, perelkiewicz2024review,huang2026survey}. Additionally, historical sources vary widely in temporal and geographic coverage, requiring careful preprocessing, metadata alignment, and expert validation to ensure reliability \citep{carey2012climate,yuan2025omnigeo}. Without dedicated resources \citep{yu2025recall}, RAG and similar systems cannot effectively retrieve evidence or perform structured reasoning for climate adaptation planning.

To address this gap, we introduce \ourmethod, the first large-scale benchmark for retrieval-augmented reasoning on historical weather archives. \ourmethod{} enables AI systems to retrieve event-specific evidence and reason over societal vulnerability and resilience indicators. It provides a systematic platform to evaluate models on handling noisy historical text, interpreting domain-specific concepts, and reasoning over complex socio-environmental narratives.

\ourmethod{} focuses on two complementary tasks: \textit{WeatherArchive-Retrieval}, which evaluates retrieval models’ ability to identify evidence-based passages corresponding to specific extreme weather events, and \textit{WeatherArchive-Assessment}, which measures LLMs’ ability to answer evidence-based queries and classify indicators of societal vulnerability and resilience using the retrieved passages. In this context, vulnerability refers to the susceptibility of communities, infrastructures, or economic systems to climate-related harm, while resilience denotes the capacity to absorb and recover from climate shocks \citep{feldmeyer2019indicators}. Understanding these dimensions from historical records is critical for identifying risk factors, designing interventions, and learning from past adaptation strategies \citep{kelman2016learning,mo2023learning,rathoure2025vulnerability}.

To support rigorous evaluation, we curate over one million OCR-parsed archival documents with dedicated preprocessing strategies, followed by expert validation and systematic quality control.
We then evaluate a range of retrieval models and state-of-the-art LLMs on three core capabilities required for climate applications: (1) processing archaic language and noisy OCR text typical of historical documents, (2) understanding domain-specific terminology and concepts, and (3) performing structured reasoning about socio-environmental relationships embedded in narratives.
Our results reveal significant limitations of current systems: dense retrieval models often fail to capture historical terminology compared to sparse methods, while LLMs frequently misinterpret vulnerability and resilience indicators. These findings highlight the need for methods that adapt to historical archival data, integrate structured domain knowledge, and reason robustly under noisy conditions.

In summary, our contributions are threefold:
\begin{enumerate}
    \item We introduce \ourmethod, which provides two evaluation tasks: \textit{WeatherArchive-Retrieval}, assessing retrieval models’ ability to extract relevant historical passages, and \textit{WeatherArchive-Assessment}, evaluating LLMs’ capacity to classify societal vulnerability and resilience indicators from archival weather narratives. 
    \item We release the first large-scale corpus of over one million historical archives, enriched through preprocessing and human curation to ensure quality, enabling both climate scientists and the broader community to leverage historical data.   
    \item We conduct comprehensive empirical analyses of state-of-the-art retrieval models and LLMs on historical climate archives, evaluating them within a fully end-to-end RAG pipeline. This exposes key limitations in handling archaic language and domain-specific terminology and provides concrete insights for building retrieval-grounded climate QA systems.
\end{enumerate}

\section{Related Work}
The urgency of addressing environmental challenges has intensified in recent decades, driven by mounting evidence of climate change, habitat degradation, and biodiversity loss \citep{rathoure2025vulnerability}. Advancing disaster preparedness requires tools that can assess vulnerabilities and resilience using realistic, context-rich cases, which urban planners and policymakers can directly act upon \citep{birkmann2015scenarios, jetz2012global}. 

Research in climate AI is limited by the scarcity of large-scale, reliable resources to capture real-world climate impacts across long temporal and geographic horizons \citep{zha2025data}. Existing data mainly target physical climate modeling or narrowly scoped contemporary text analysis, leaving historical, case-based impact records largely untapped. For instance, ClimateIE \citep{pan2025climateie} provides 500 annotated climate publications aligned with the GCMD+ taxonomy but focuses on technical entities such as observational variables rather than societal consequences of extreme weather.   
Our work addresses this gap by constructing a large-scale benchmark of real-world climate impact narratives. The dataset is curated from archival sources covering diverse events over extended periods, ensuring \textbf{reliability} through careful annotation and \textbf{reusability} for information retrieval, text mining, and case-based analysis tasks. 

\section{\ourmethod}
Our goal with \ourmethod{} is to provide a realistic benchmark for evaluating current retrieval and reasoning capabilities in the context of climate- and weather-related archival texts. In particular, we focus on the dual challenges of (i) constructing a high-quality corpus from archival news segments and (ii) defining retrieval and generation tasks that capture the practical needs of climate researchers. This section details our corpus collection pipeline and task formulation. 

\subsection{Corpus Collection}
\label{subsec:dataset} 
LLMs are generally pre-trained on large-scale internet corpora, which frequently include fake and unreliable content \citep{roy2021fake}. In contrast, archival news segments provide a unique and valuable information source, as copyright restrictions typically exclude them from LLM pretraining data. Unlike standardized meteorological datasets, archival news segments provide rich, narrative descriptions of weather-related disruptions and community-level adaptation successes \citep{sieber2024identifying}. These archives also capture public voices and societal perspectives that would be prohibitively expensive to collect today, yet public perceptions remain documented in historical records \citep{thurstan2016oral}. 
Thus, our corpus offers contextualized insights that complement traditional climate data. For climate scientists seeking to understand long-term patterns of societal vulnerability and resilience, these narrative-rich sources provide invaluable evidence of how communities have historically experienced, interpreted, and responded to weather-related challenges.

\begin{table}[t]
\centering
\small
\caption{OCR correction quality: \textsc{GPT-4o} vs. Human Annotations ($n=50$).}
\label{tab:ocr_quality}
\begin{tabular}{@{}lcccc@{}}
\toprule
\textbf{Metric} & \textbf{1-gram} & \textbf{2-gram} & \textbf{3-gram} & \textbf{L} \\
\midrule
\textbf{BLEU}   & 0.911 & 0.853 & 0.817 & ---   \\
\textbf{ROUGE}  & 0.947 & 0.919 & ---   & 0.943 \\
\bottomrule
\end{tabular}
\vspace{-10pt}
\end{table}

Our corpus construction emphasizes both scale and reliability. Sourced from a proprietary archive institution, we collected two 20-year tranches of news archives from an organization in Southern Quebec, a region representative of broader Northeastern American weather patterns: one covering a contemporary period (1995–2014) and one covering a historical period (1880–1899). The archival news articles were digitized with OCR and subsequently cleaned using \textsc{GPT-4o}, following the post-OCR correction method of ~\citet{zhang2024post}. To validate this pipeline, we compared GPT-4o's corrections against expert human transcriptions for a random sample of 50 articles. As shown in Table~\ref{tab:ocr_quality}, the high BLEU and ROUGE scores (e.g., ROUGE-L of 0.943) confirm that the automated process effectively removes OCR noise while preserving the semantic integrity required for climate reasoning. Although OCR noise is a known issue in archival processing, retaining it would distort our evaluation of climate comprehension and cross-document reasoning, which are the core goals of our benchmark. 
Unlike OCR-focused datasets such as OHRBench \citep{liu2024ocrbench} that vary noise levels to study error cascades, our benchmark intentionally provides OCR-corrected text to isolate climate-specific retrieval and societal-impact reasoning.
We then segmented the archival news articles into overlapping archival news segments using a sliding-window approach, followed by the method proposed by~\citet{sun2023chatgpt}, allowing each segment to preserve sufficient semantic context while satisfying token-length constraints. The resulting dataset comprises 1,035,862 news segments, each standardized to approximately 256 tokens, which we used for the \textit{WeatherArchive-Retrieval} task creation.

\subsection{Task definition}
\ourmethod{} incorporates two complementary tasks designed to mirror the workflow of climate scientists. \textit{WeatherArchive-Retrieval} tests models' ability to locate relevant historical evidence. The other is \textit{WeatherArchive-Assessment}, which evaluates their capacity to interpret complex socio-environmental relationships within an archival report of an extreme weather event.

\label{task 1}
\begin{figure}[htbp]
    \centering
    \includegraphics[width=\columnwidth]{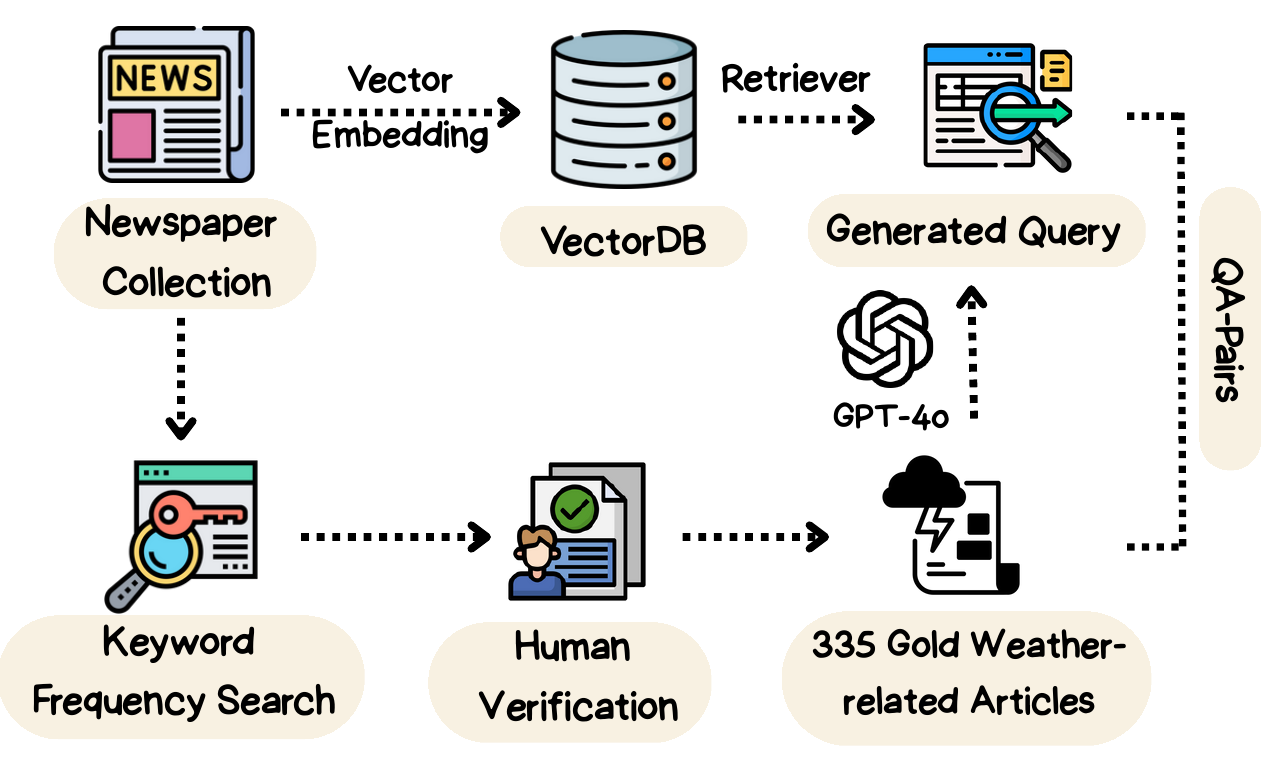}
    \caption{The construction pipeline of the retrieval task in weather archive collections.} 
    \label{fig-retrieve-stage}
\end{figure}

\begin{figure*}[!htb]
    \centering
    \includegraphics[width=\textwidth]{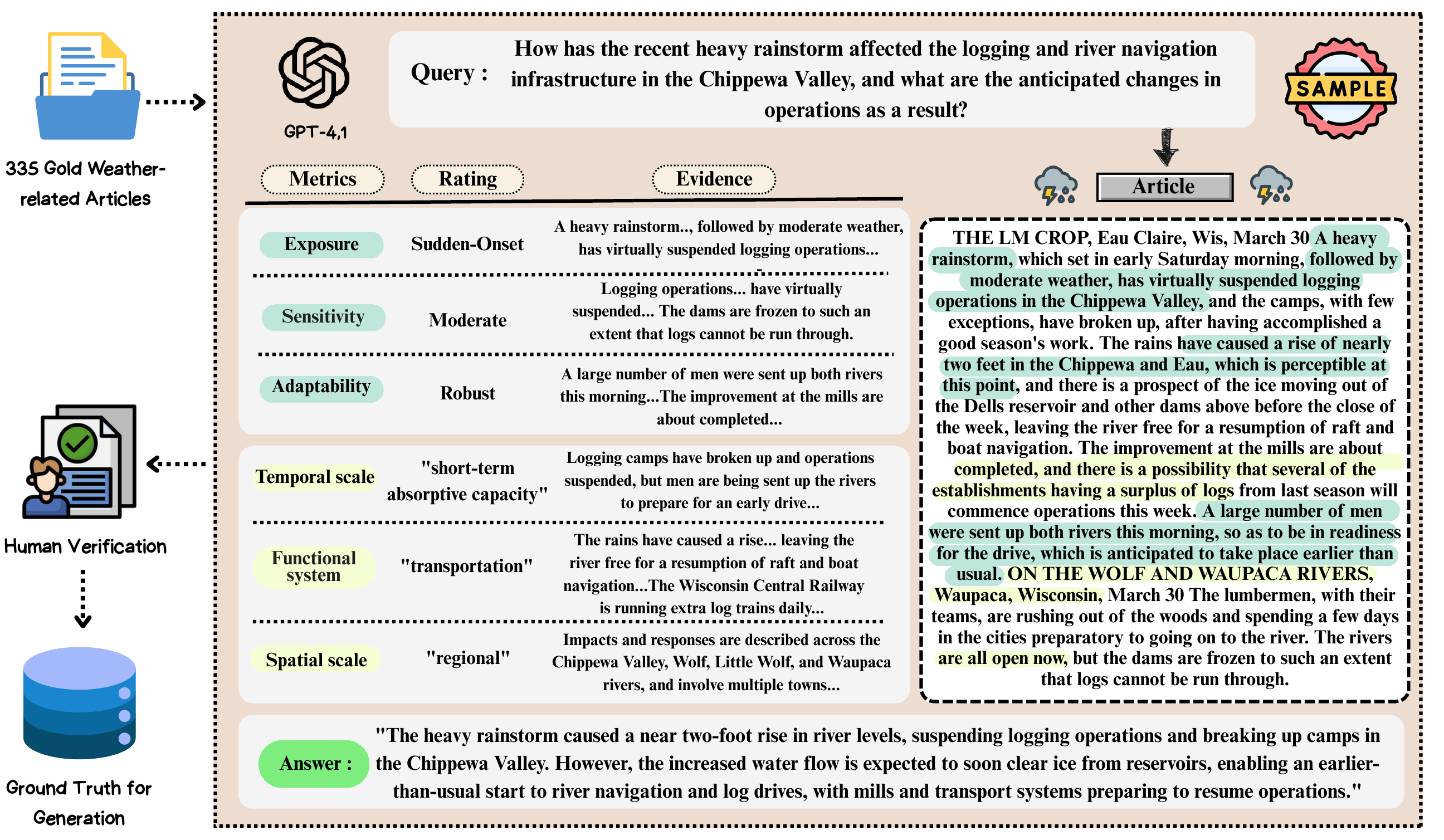}

    \label{fig:WeatherArchive-Assessment}
    \caption{\textit{WeatherArchive-Assessment} - the construction pipeline of assessment task on societal vulnerability and resilience. \textsc{GPT-4.1} evaluates retrieved archival news segments across multiple criteria, with human verification ensuring quality before generating ground truth answers. This sample case shows the assessment of rainstorm impacts.}
\end{figure*}

\subsubsection{WeatherArchive-Retrieval}
In scientific domains such as climate analysis, scientists often rely on precedents embedded in long historical archives \citep{Herrera2003-jx, Slonosky2020-as, Sieber2022-tj}. A well-designed retrieval task (Figure ~\ref{fig-retrieve-stage}) is essential, as it evaluates a model’s ability to identify contextually relevant and temporally appropriate information while providing a reliable foundation for subsequent answer generation.

To construct the benchmark, we first ranked 1,035,862 archival news segments by the frequency of keywords related to disruptive weather events. From this ranking, we selected the top 525 segments, which were then manually reviewed by domain experts to identify those providing complete evidential support for end-to-end question answering. After curation, 335 high-quality validated segments were retained. For each passage, we generated domain-specific queries using \textsc{GPT-4o}. These queries were designed to emulate real-world research intents, resulting in a realistic retrieval benchmark composed of query–answer pairs.

The difficulty of this task stems from the nature of the segments extracted from historical archives. Unlike contemporary datasets, news archives use domain-specific terminology that has shifted over time (e.g., outdated expressions for storms or floods), which makes relevance judgments nontrivial. Moreover, archival news articles frequently embed descriptions of weather impacts within broader narratives or unrelated sections such as advertisements, which introduces additional noise into the retrieval process. By grounding evaluation in such historically situated and noisy data, \textit{WeatherArchive-Retrieval} establishes a challenging yet realistic testbed for assessing the robustness of retrieval models and systems.

\subsubsection{WeatherArchive-Assessment}
\label{weather understanding}
To effectively support climate scientists in disaster preparedness, language models must go beyond retrieving relevant news segments and demonstrate the ability to interpret societal vulnerability and resilience as documented in historical texts. To this end, we design an evaluation framework to assess a model’s ability to reason about climate impacts across multiple levels, drawing on established approaches from vulnerability and adaptation research \citep{feldmeyer2019indicators}. The framework comprises two complementary subtasks: (i) classification of societal vulnerability and resilience indicators, and (ii) open-ended question answering to assess model generalization on climate impact analysis. To clarify the task setup, we summarize the dataset construction here: historical newspaper narratives are paired with daily weather records from the same location and period. Each example includes a climate-related query and retrieved passages reflecting the documented weather impact. Archival news articles are digitized, cleaned, and aligned with meteorological data to ensure both sources describe the same event. Models are required to link narrative evidence with the associated weather record to produce an answer, which is evaluated against aligned labels. Our experimental pipeline follows a retrieve-then-answer paradigm: the retriever selects relevant archival segments using the query exactly as written, and the QA model generates its response solely from the retrieved passages. The \textit{WeatherArchive-Assessment} task contains 335 examples, with a one-to-one mapping between historical weather reports and queries, yielding query–answer pairs generated using \textsc{GPT-4.1} with structured prompting.

\begin{table*}[ht]
\centering
\caption{F1 evaluation results (in percentages) for vulnerability and resilience indicator classification on \textit{WeatherArchive-Assessment} across diverse LLMs. \textbf{Bold} and \underline{\textit{underline}} indicate the best and second-best results.}
\label{vulnerability&resilience results-F1}
\resizebox{\textwidth}{!}{%

\renewcommand{\arraystretch}{0.85}
\begin{tabular}{lccccccc}
\toprule\toprule
& \multicolumn{3}{c}{Vulnerability (\%)} & \multicolumn{3}{c}{Resilience (\%)} & \multirow{2}{*}{Average (\%)} \\
\cmidrule(lr){2-4} \cmidrule(lr){5-7}
Model & Exposure & Sensitivity & Adaptability & Temporal & Functional & Spatial & \\
\midrule\midrule
\textsc{gpt-4o}            
    & 64.6 & 52.8 & 58.0 & 62.3 & \textbf{64.5} & 51.8 & \cellcolor{gray!15}59.0 \\
\textsc{gpt-3.5-turbo}     
    & 63.6 & 46.6 & 46.5 & 64.3 & 34.2 & 39.5 & \cellcolor{gray!15}49.1 \\
\textsc{claude-opus-4-1}   
    & \underline{78.3} & \underline{67.6} & 67.5 & \textbf{84.6} & 62.5 & \textbf{61.4} & \cellcolor{gray!15}\textbf{70.3} \\
\textsc{claude-sonnet-4}   
    & 77.2 & \textbf{73.8} & 59.7 & 65.2 & \underline{63.5} & 60.3 & \cellcolor{gray!15}\underline{66.6} \\
\textsc{gemini-2.5-pro}    
    & 76.6 & 62.0 & 57.1 & 75.6 & 62.5 & \underline{61.3} & \cellcolor{gray!15}65.8 \\
\midrule
\textsc{DeepSeek-V3-671B}  
    & \textbf{79.8} & 49.5 & \textbf{70.9} & \underline{76.0} & 61.3 & 60.8 & \cellcolor{gray!15}66.4 \\
\textsc{Mixtral-8x7B-IT}   
    & 27.3 & 21.4 & 24.1 & 32.2 & 21.4 & 32.6 & \cellcolor{gray!15}26.5 \\
\textsc{Ministral-8B-IT}   
    & 43.7 & 18.8 & 24.6 & 45.8 & 41.9 & 37.0 & \cellcolor{gray!15}35.3 \\
\textsc{Qwen3-30B-IT}      
    & 65.8 & 44.4 & 30.0 & 73.0 & 34.2 & 36.4 & \cellcolor{gray!15}47.8 \\
\textsc{Qwen3-4B-IT}       
    & 32.0 & 27.5 & 18.4 & 49.6 & \textbf{64.5} & 28.5 & \cellcolor{gray!15}36.8 \\
\textsc{Qwen2.5-72B-IT}    
    & 74.4 & 43.4 & \underline{67.6} & 73.5 & 49.8 & 51.5 & \cellcolor{gray!15}60.0 \\
\textsc{Qwen2.5-7B-IT}     
    & 33.8 & 9.1 & 22.5 & 33.0 & 30.8 & 32.9 & \cellcolor{gray!15}27.0 \\
\textsc{Llama-3.3-70B-IT}  
    & 36.7 & 42.9 & 24.4 & 48.1 & 53.1 & 35.5 & \cellcolor{gray!15}40.1 \\
\textsc{Llama-3-8B-IT}     
    & 24.3 & 19.8 & 18.4 & 19.4 & 29.0 & 28.6 & \cellcolor{gray!15}23.3 \\
\midrule
\rowcolor{gray!15}
Average 
    & 54.7 & 40.6 & 41.5 & 56.2 & 46.5 & 42.8 & \cellcolor{gray!30}47.0 \\
\bottomrule\bottomrule
\end{tabular}}
\end{table*}

\paragraph{Societal Vulnerability.}  
Vulnerability is widely conceptualized as a function of exposure, sensitivity, and adaptability \citep{o2004vulnerable}. We operationalize this framework by prompting models to assign descriptive levels to each component. 
Specifically, \textit{exposure} characterizes the type of climate or weather hazard, distinguishing between sudden-onset shocks (e.g., storms, floods), slow-onset stresses (e.g., prolonged droughts, sea-level rise), and compound events involving multiple interacting hazards. \textit{Sensitivity} evaluates how strongly the system is affected by such hazards, ranging from critical dependence on vulnerable resources to relative insulation from disruption. \textit{Adaptability} captures the ability of the system to respond and recover, spanning robust governance and infrastructure to fragile conditions with little or no coping capacity.

This classification-based evaluation examines whether models can move beyond surface-level text interpretation toward structured reasoning about vulnerability, which is essential for anticipating climate risks \citep{linnenluecke2012extreme}. In practice, \textit{exposure} and \textit{adaptability} are often signalled by explicit indicators \citep{brooks2005determinants} such as infrastructure damage or recovery measures, which evaluate LLMs' capacity to capture through climate factual extraction. \textit{Sensitivity} is more challenging, as it requires climate reasoning \citep{montoya2023socio} about governance quality, institutional strength, or social capital, factors that are seldom directly expressed in segments. By incorporating both explicit and implicit aspects of vulnerability, our framework provides a rigorous test of whether models can integrate factual evidence with contextual inference.

\paragraph{Societal Resilience.}  
Resilience is evaluated using indicators proposed by \citet{feldmeyer2019indicators}, which emphasize adaptation processes across three scales. On the \textit{temporal scale}, models must distinguish between short-term absorptive capacity (e.g., emergency response), medium-term adaptability (e.g., policy or infrastructure adjustments), and long-term transformative capacity (e.g., systemic redesign). On the \textit{functional system scale}, models identify which systems are affected, including health, energy, food, water, transportation, and information, highlighting their interdependence in shaping preparedness. Lastly, on the \textit{spatial scale}, models assess resilience across levels (e.g., local, community, regional, national), capturing variation in adaptability across contexts. Through the experts' annotation process, we are informed that temporal indicators are often easier to identify since newspapers tend to report immediate damages and responses explicitly, whereas functional and spatial dimensions are more challenging since they require models to infer systemic interactions and contextual variation that are rarely stated explicitly in news archives. By formulating these criteria into multiple-choice questions, we evaluate whether models can recognize structured indicators of resilience within noisy archival narratives.


\subsubsection{Oracle Quality and Validation}
To validate the LLM-generated queries and oracles, four independent climate librarian experts annotate shared subsets of 60 segments. This evaluation yields a substantial inter-annotator agreement of $\kappa_{\text{Fleiss}} = 0.76$, and \textsc{GPT-4.1} achieves an accuracy of $0.82$ with respect to the adjudicated ground truth. To strengthen this validation, we adopt the statistical framework proposed by \citet{calderon2025alternative}, with a winning rate of $\omega = 0.75$ (where $\omega > 0.5$ indicates higher agreement with the reference annotations than the average individual annotator). This framing emphasizes alignment with the annotation protocol rather than any normative comparison to human expertise.

\section{Experimental Setup}
\subsection{Evaluation Metrics}
\label{metrics}
\paragraph{WeatherArchive-Retrieval.}  We evaluate retrieval performance with the commonly used metrics, including Recall@$k$, MRR@$k$, and nDCG@$k$ for $k \in \{3,10,50,100\}$.

\paragraph{WeatherArchive-Assessment.} 
The downstream benchmark evaluates climate-related reasoning using expert-validated references along two dimensions. Vulnerability and resilience indicator classification requires models to identify and categorize societal factors from historical weather narratives, evaluated using precision, recall, and F1. Historical climate question answering assesses models’ ability to generate evidence-based responses from retrieved archival passages, measured using BLEU, ROUGE, BERTScore, and token-level F1 against expert-authored answers. To evaluate reasoning beyond surface-level similarity, we additionally employ LLM-based judgment with \textsc{GPT-4.1}, which compares model outputs to oracle answers and marks responses correct if they match or subsume the reference without factual errors.

\paragraph{Question Answering (QA) in Climate AI}
\label{QA}
Each climate-related query is first processed by the retrieval component, which returns the top-3 news segments as the model’s sole evidence. This setup enables end-to-end evaluation of RAG systems, where generation quality directly depends on retrieval quality, reflecting real-world deployment scenarios. When retrieved evidence is insufficient, models are expected to acknowledge uncertainty rather than hallucinate answers. Gold-standard answers are used only for evaluation, allowing us to assess both retrieval effectiveness and the LLM’s ability to reason under imperfect evidence. While vulnerability and resilience classification evaluates structured evidence extraction, free-form QA tests models’ ability to synthesize dispersed archival information and articulate climate impacts.

\begin{figure*}[!htb]
    \centering    \includegraphics[width=0.95\textwidth]{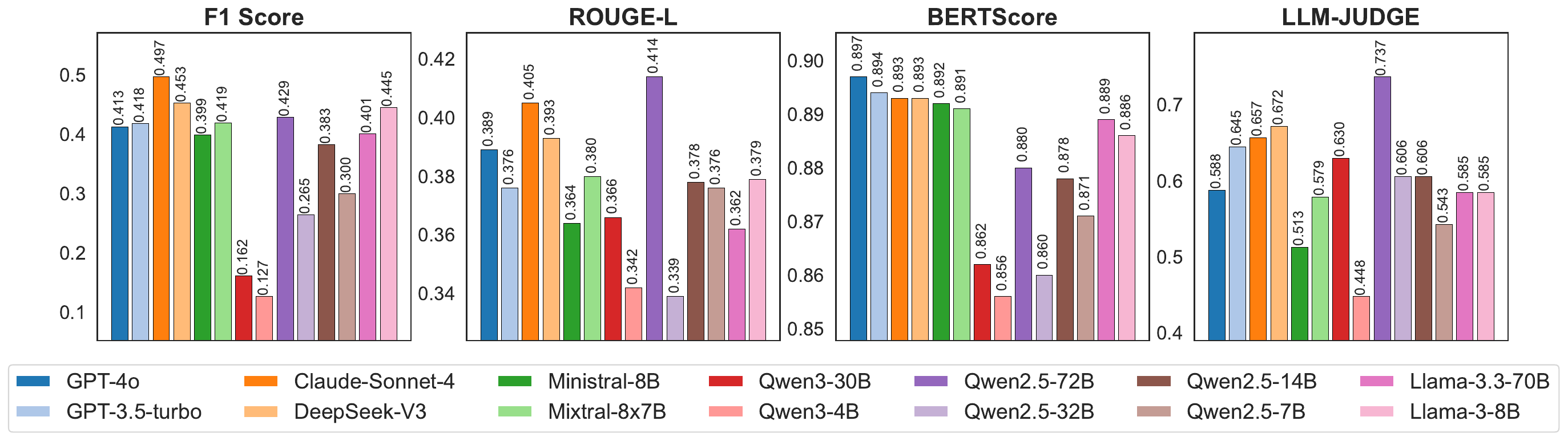}
    \caption{Performance comparison of LLMs on free-form QA task across various metrics.}
    \label{QA results}
\end{figure*}

\subsection{Evaluated Models}
\paragraph{Retrieval Models.} We benchmark retrieval models on the archival collections, including three categories: (i) sparse lexical models: BM25 \citep{robertson2009probabilistic} (ii) dense embedding models: ANCE \citep{xiong2020approximate}, SBERT \citep{reimers2019sentence}, and large proprietary embeddings, including OpenAI’s text-embedding-3-large and text-embedding-ada-002 \citep{neelakantan2022text}, Gemini’s text-embedding \citep{lee2025gemini}, IBM’s Granite Embedding \citep{awasthy2025granite}, and Snowflake’s Arctic-Embed \citep{yu2024arctic} and (iii) re-ranking models: cross-encoders applied on BM25 candidates (BM25+CE) with a MiniLM-based reranker \citep{wang2020minilm}.

\paragraph{Language Models.} We consider a diverse suite of open-source and proprietary LLMs with various parameter scales. Open-source models include Qwen-2.5 (7B–72B), Qwen-3 (4B, 30B), LLaMA-3 (8B, 70B), Mistral-8B and Ministral-8×7B. These families capture scaling effects, efficiency–performance trade-offs, and robustness to long or noisy text. We also include DeepSeek-V3-671B, which targets efficient scaling and adaptability. Proprietary models include GPT (3.5-turbo, 4o), Claude (opus-4-1, sonnet-4) and Gemini-2.5-pro, which are widely used in applied pipelines, offering strong reasoning and summarization capabilities. All models are instruction-tuned versions, denoted as ``IT''.

\section{Experimental Results}

\subsection{WeatherArchive-Retrieval Evaluation}

\begin{table}[t]
\centering
\caption{{Retrieval performance (in percentages)} on \textit{WeatherArchive-Retrieval}. 
\textbf{Bold} and \underline{\textit{underline}} indicate the best and the second-best performance.}
\label{tab:recall_ndcg_results}

\small
\setlength{\tabcolsep}{4pt}

\begin{tabular}{@{}llcc|cc@{}}
\toprule\toprule
& & \multicolumn{2}{c}{\textbf{Recall (\%)}} & \multicolumn{2}{c}{\textbf{nDCG (\%)}} \\
\cmidrule(lr){3-4} \cmidrule(lr){5-6}
\textbf{Category} & \textbf{Model} & \textbf{@3} & \textbf{@10} & \textbf{@3} & \textbf{@10} \\
\midrule\midrule
\textbf{Sparse} 
    & \textsc{bm25}   & \underline{58.5} & 67.8 & \underline{49.7} & 53.2 \\
\midrule
\textbf{Dense} 
    & \textsc{sbert}  & 29.0 & 40.0 & 22.8 & 26.8 \\
    & \textsc{ance}   & 34.0 & 52.2 & 27.3 & 33.8 \\
    & \textsc{arctic} & 53.4 & 67.5 & 44.3 & 49.4 \\
    & \textsc{granite} & 54.6 & 71.9 & 44.8 & 51.2 \\
    & \textsc{openai-3-large} & 48.1 & 65.1 & 40.0 & 46.1 \\
    & \textsc{openai-ada-002} & 51.0 & 70.2 & 42.1 & 49.2 \\
    & \textsc{gemini-embedding-001} 
        & 57.3 & \underline{74.9} & 47.9 & \underline{54.3} \\
\midrule
\textbf{Re-Ranking} 
    & \textsc{BM25+CE} & \textbf{63.9} & \textbf{76.1} & \textbf{53.2} & \textbf{57.9} \\
\bottomrule\bottomrule
\end{tabular}
\normalsize
\vspace{-2ex}
\end{table}

\noindent \paragraph{Sparse Retrieval Models Achieve Strong Top-rank Relevance on Climate Archives.}
As shown in Table~\ref{tab:recall_ndcg_results}, BM25 variants continue to perform strongly, often matching or surpassing dense alternatives in ranking quality at top $k$.
The effectiveness of BM25 might be related to the nature of climate-related queries, which usually contain technical terminology and domain-specific collocations (e.g., ``flood damage,'' ``hurricane casualties,'' ``crop failure due to drought''). 
In such cases, exact lexical matching is critical as sparse methods are able to capture these specialized terms directly, whereas dense representations may blur over distinctions or concepts. 
For instance, a query about ``storm surge fatalities'' would benefit from precise overlap with news segments containing the same terminology, whereas a dense retriever might incorrectly emphasize semantically related but distinct expressions such as ``storm warnings'' or ``storm intensity''. These findings highlight the importance of sparse methods in scientific and technical domains where specialized vocabulary governs relevance.

\noindent \paragraph{Re-ranking Could Boost Performance.}
With the effective sparse methods, further deploying a re-ranker brings better performance.
In this setup, BM25 provides high lexical coverage at the candidate generation stage, and the re-ranker ranks the top candidates by modelling fine-grained query–document interactions. 
Empirically, the results show hybrid model (BM25+CE) consistently outperforms both pure sparse and pure dense baselines within the top-ranked results (e.g., top 3-10 segments), which are most critical for downstream QA. This indicates that re-ranking models with a baseline yield more robust performance for climate-related retrieval.

\subsection{WeatherArchive-Assessment Evaluation}

\paragraph{Factual Extraction vs. Socio-environmental Reasoning.}
Consistent with prior scaling-law findings \citep{kaplan2020scaling}, larger models generally achieve better zero-shot performance. As shown in Table~\ref{vulnerability&resilience results-F1}, models perform well on indicators that rely on explicit factual extraction, such as infrastructure damage and recovery actions. In contrast, sensitivity and resilience indicators require deeper reasoning about how weather shocks affect interdependent socio-environmental systems \citep{morss2011improving}. While models show relatively strong performance on temporal dimensions, reflecting their ability to identify immediate response capacities, performance degrades on functional and spatial dimensions. Even advanced models struggle with cross-system dependencies and multi-scale coordination, often over-predicting system-level impacts or overlooking localized effects. These results highlight persistent limitations in LLMs’ ability to reason about distributed, scale-dependent climate impacts, underscoring the continued need for human expertise in vulnerability assessment.

\paragraph{LLMs Struggle with Socio-environmental System Effects.}
Societal resilience indicator classifications require recognizing direct damages from disruptive weather events and reasoning about how shocks propagate across geographic scales and interdependent systems.
As shown in Table~\ref{vulnerability&resilience results-F1}, models achieve relatively strong performance on temporal dimensions with a score of 56.2\% on average, with \textsc{Claude-Opus-4-1} and \textsc{DeepSeek-V3-671B} reliably identifying immediate response capacities. 
However, performance degrades on functional and spatial dimensions, where even sophisticated models struggle to assess cross-system dependencies (e.g., over-predicting ``transportation'' or ``information'') and multi-scale coordination (e.g., overlooking ``local''). 
Impacts are distributed unevenly across systems and exhibit inherently scale-dependent propagation dynamics. This pattern reveals limitations as models perform well at identifying direct impacts, yet are limited in reasoning over complex socio-environmental interdependencies that mediate systemic resilience. This highlights that multi-scale vulnerability assessment still requires human expertise.

\textit{From Retrieval to Reasoning: LLM Performance on Climate-specific QA.} We evaluate retrieval-augmented LLMs on their ability to synthesize historical climate news into coherent, domain-specific answers. As shown in Figure~\ref{QA results}, retrieval quality has a substantial impact on downstream QA performance, with relevant context consistently improving generation quality across all metrics compared to the no-context setting. Lexical and semantic evaluations reveal complementary strengths, indicating that accurate climate-specific QA requires both precise evidence grounding and higher-level reasoning. Overall, while larger models can effectively integrate retrieved information, generating scientifically accurate free-form answers from historical archives remains challenging, highlighting persistent gaps in climate-specific reasoning.





\section{Conclusion}
\ourmethod{} establishes the first large-scale benchmark for evaluating the full RAG pipeline on historical weather archives. By releasing a dataset of over one million archival news segments, it enables climate scientists and the broader community to leverage historical data at scale. With well-defined downstream tasks and evaluation protocols, the benchmark rigorously tests both retrieval models and LLMs. In doing so, it transforms underutilized archival narratives into a standardized resource for advancing climate-focused AI.

Our analyses reveal that hybrid retrieval approaches outperform dense methods on historical vocabulary, while even proprietary LLMs remain limited in reasoning about vulnerabilities and socio-environmental dynamics. Future research should address two identified challenges: (1) enhancing retrieval methods to better handle historical vocabulary and narrative structures, and (2) improving models' ability to reason about complex socio-environmental systems beyond surface-level factual extraction. By offering a standardized evaluation resource, \ourmethod{} lays the groundwork for future research toward AI systems that can translate historical climate experience into actionable intelligence for adaptation and disaster preparedness.

\section*{Ethics Statement}
The \ourmethod{} is built from a collection of digitized historical newspapers provided through collaboration with the McGill University Library, the Bibliothèque nationale du Québec, and the Montreal Gazette. This organization retains the copyright of the archival news articles, but has granted permission to publish the curated benchmark in support of the climate research community. We additionally acknowledge that using GPT-based post-OCR correction may introduce model-driven biases, which we treat as an important consideration for the integrity of the task.

Although the majority of extreme weather events in our dataset are recorded in North America, the accounts capture how societies experienced and responded to climate hazards. These records provide broadly relevant insights into resilience strategies and adaptation planning that extend beyond their original geographical context. In addition, contributions from crowd-sourcing may be influenced by geodemographic factors, which introduces variation but also enriches the dataset \citep{hendrycks2020aligning}. As such, the benchmark reflects diverse societal perspectives on climate impacts and responses, making it a valuable resource for studying adaptation strategies across societal contexts.

\section*{Acknowledgment}
Our primary data source is a corpus of three digitized newspapers (La Presse, La Patrie, and Montreal Gazette), obtained through collaboration with the McGill University Library and Archives and the Bibliothèque nationale du Québec. We would like to thank DRAW McGill for their guidance throughout this project, especially Dr. Victoria Slonosky.

We are also deeply grateful for the support provided by OpenAI Grants and RBC Borealis AI. This research was further supported by the McCAIS grant funding and CEIMIA (Centre d'expertise international de Montréal en intelligence artificielle), which funds the research, development, and innovation in artificial intelligence.


\bibliographystyle{ACM-Reference-Format}
\balance
\bibliography{wxRAG}

\end{document}